\documentclass{article}
\usepackage{spconf,amsmath,graphicx}
\usepackage{threeparttable}
\usepackage{physics}

\title{Two-Stage Framework for Seasonal Time Series Forecasting}
\name{Qingyang Xu\textsuperscript{\rm 1,2}, Qingsong Wen\textsuperscript{\rm 3}, Liang Sun\textsuperscript{\rm 3}}
\address{
\textsuperscript{\rm 1}MIT Operations Research Center, Cambridge, USA\\
\textsuperscript{\rm 2}MIT Laboratory for Financial Engineering, Cambridge, USA\\
\textsuperscript{\rm 3}Machine Intelligence Technology, Alibaba Group, 
Bellevue, USA\\
}

\begin{document}
%
\maketitle

\begin{abstract}
Seasonal time series Forecasting remains a challenging problem due to the long-term dependency from seasonality. In this paper, we propose a two-stage framework to forecast univariate seasonal time series. The first stage explicitly learns the long-range time series structure in a time window beyond the forecast horizon. By incorporating the learned long-range structure, the second stage can enhance the prediction accuracy in the forecast horizon. In both stages, we integrate the auto-regressive model with neural networks to capture both linear and non-linear characteristics in time series. Our framework achieves state-of-the-art performance on M4 Competition Hourly datasets. In particular, we show that incorporating the intermediate results generated in the first stage to existing forecast models can effectively enhance their prediction performance. 
\end{abstract}
\begin{keywords}
Time series, seasonality, forecasting, deep learning
\end{keywords}

\section{Introduction}


Time series signal processing and mining have been widely researched in a variety of fields~\cite{esling2012time,angelosante2011sparse,wen2020robust}. 
Among them, seasonal or periodic time series is commonly observed in many real-world applications~\cite{banerjee2018time,FastRobustSTL_wen2020}. Seasonality generally refers to the repeated pattern with long-term dependency which affects the time series signal. 
Forecasting univariate time series with seasonality has many important real-world applications. One major scenario is the proactive auto-scaling of computing resources~\cite{bauer2020time,taft2018p}. Empirically, the demand for computing resources often exhibits strong periodic patterns (sometimes multi-periodic). To satisfy resource demands for all users while minimizing the amount of unused resources, one must accurately predict the future demand in a forecast horizon of length $h>1$ and adjust the resource allocation dynamically. Inefficient allocations of computing resources may incur significant costs for the company.

Traditional linear forecast methods for seasonal univariate time series such as SARIMA and Error-Trend-Seasonality (ETS)~\cite{hyndman2008forecasting} are computationally efficient and interpretable but have limited prediction accuracy since they do not capture the complex nonlinear dependence in real-world time series. Seasonal Periodic Autoregression (SPAR)~\cite{SPAR08} is a strong benchmark to forecast seasonal time series one step ahead (i.e. $h=1$) but its accuracy quickly deteriorates when we predict longer forecast horizons (e.g. $h=T/2$) by using predicted value of $\hat{x}_{t+1}$ as input to predict $x_{t+2}$. Recently, deep learning model like RESTFul~\cite{wu2018restful} can achieve state-of-the-art results but its encoding mechanism still cannot effectively capture long-range structure in complex seasonal time series.

Given these challenges, we propose a novel two-stage framework. Our method simplifies the pre-training stage of the standard self-supervised learning (SSL) framework~\cite{liu2020selfsupervised} by explicitly learning the time series $x_t$  in a time window after the forecast horizon. The prediction stage incorporates the long-range time series structure to enhance the prediction accuracy in the forecast horizon. We find that the two-stage framework utilizes the seasonal structure of $x_t$ more effectively and is highly computationally efficient and interpretable compared to other SSL-based methods.

\section{TWO-STAGE forecasting framework}

\subsection{Model Description}
Given a univariate time series $x_t$ with period length $T$, our goal is to predict its values in the {forecast horizon} $\vb{x}_f=(x_{t+1}, x_{t+2}, \cdots, x_{t+h})$ with $h>1$ using its historical values $\vb{x}_{his}=(x_{t-L+1},…,x_{t-1},x_t)$. We adopt the RobustPeriod algorithm~\cite{wen2020robustperiod} to estimate the period length $T$.
For time series with multiple periods (e.g. an hourly time series may have periodic patterns on both daily and weekly basis), we define $T$ as the shortest period length.

\subsection{Proposed Framework}

In this paper, we propose a novel framework to forecast univariate seasonal time series motivated by self-supervised learning (SSL). 
SSL can in general be regarded as a two-stage framework~\cite{liu2020selfsupervised}. Stage 1 (“pre-training”) extracts a latent feature representation from a large amount of unlabeled data by recovering part of the input data (often selected with random masking) from its visible parts. Stage 2 (“fine-tuning”) transfers these hidden representations to a specific downstream task, often trained with limited labeled data. In recent years, SSL has enjoyed tremendous success in sequential data modeling \cite{sermanet2018time,baevski2019vq, wang2019reinforced}. One of the most notable SSL-based models is BERT \cite{devlin2019bert}, which achieved the state-of-the art performance in a variety of NLP tasks from contextual inference to next sentence prediction.


However, in univariate time series forecast, SSL has not achieved similar success so far. We argue that this is due to two challenges in this domain.  First, since the time series $x_t$ is univariate, it is intrinsically a “small data” problem and the pre-training step via random masking is both computationally inefficient and likely to result in overfitting the encoder. In addition, for seasonal time series, the random masking pre-training does not effectively utilize the seasonal structure (such as period length $T$). 

To overcome the limitation of SSL for time series analysis, we propose a two-stage forecasting framework specifically for  univariate seasonal time series. Stage 1 of the proposed framework learns a mapping $f_1^*$ from historical time series $\vb{x}_{his}$ to predict the {future horizon} $\vb{x}_{fut}=(x_{t+h+1},x_{t+h+2}, \cdots, x_{t+h+H})$ of length $H>1$ after the forecast horizon. The length of future horizon $H$ is a critical hyperparameter which needs to be adjusted for different values of $T$ and $h$. Stage 2 of the proposed framework consists of two steps: training and prediction. The training step learns a mapping $f_2^*$ from both $\vb{x}_{his}$ and the true values of future horizon $\vb{x}_{fut}$ to predict forecast horizon $\vb{x}_f=f_2^* (x_{his},x_{fut})$. Finally, in the prediction step, we first predict the future horizon via Stage 1 mapping $\hat{\vb{x}}_{fut} = f_1^*(\vb{x}_{his})$ and then use both $\vb{x}_{his}$ and $\hat{\vb{x}}_{fut}$ to predict the forecast horizon, i.e.
\begin{equation}
\hat{\vb{x}}_f = f_2^*(\vb{x}_{his}, \hat{\vb{x}}_{fut}) =  f_2^*(\vb{x}_{his}, f_1^*(\vb{x}_{his}))
\end{equation}
The workflow of the two-stage framework is given in Fig. 1 and a schematic illustration of the model is given in Fig. 2.

\begin{figure}[!t]
\centering
    \includegraphics[width=1\linewidth]{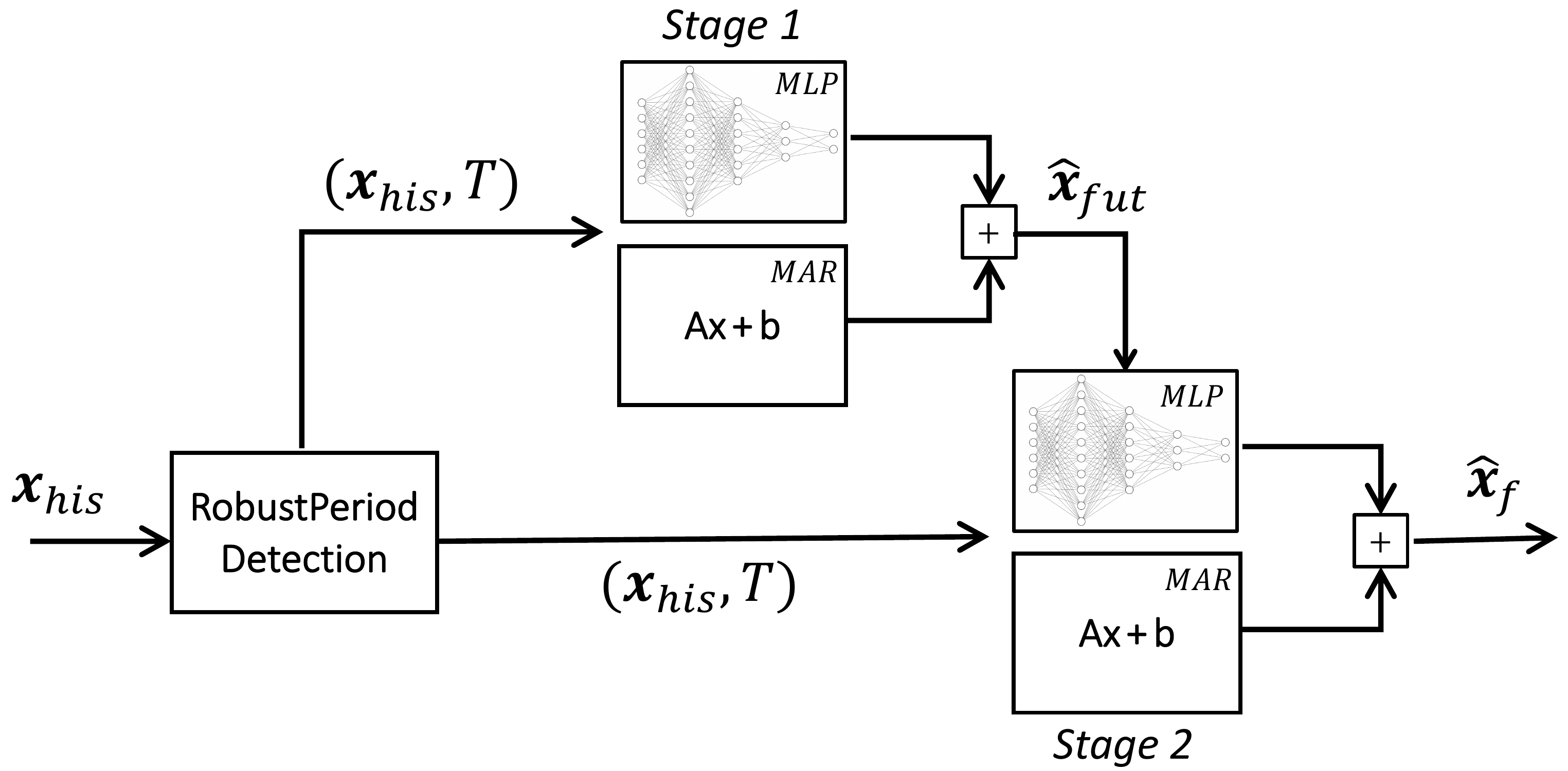}
    \caption{Diagram of the proposed two-stage framework for seasonal time series forecasting.}
\label{two_stage_scheme}
\end{figure}

\begin{figure}[!t]
\centering
\includegraphics[width=0.6\linewidth]{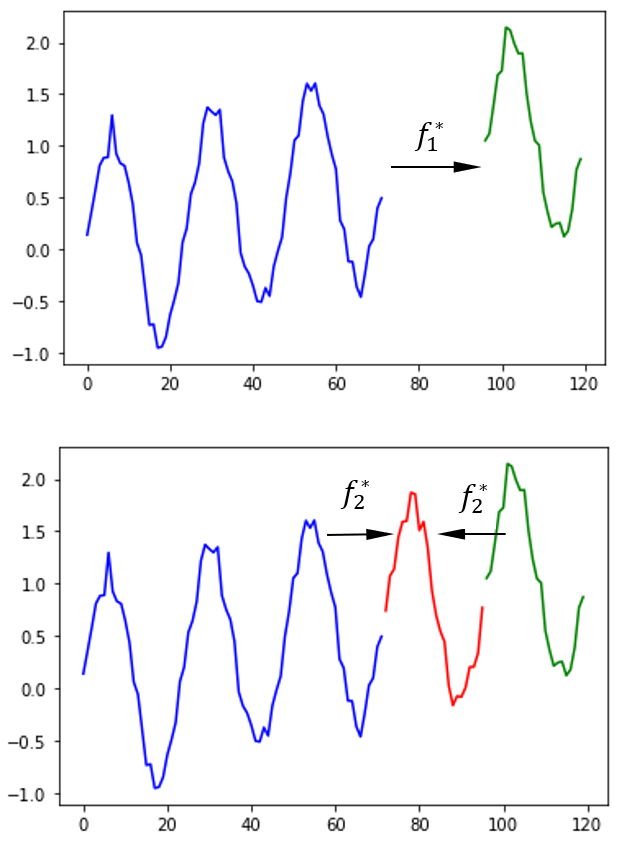}
\caption{Schematic illustration of two-stage model. Upper figure: Stage 1 learns future horizon. Lower figure: Stage 2 uses both historical and future time series to predict the forecast horizon. (Blue: historical time series $\vb{x}_{his}$. Red: forecast horizon $\vb{x}_f$. Green: future horizon $\vb{x}_{fut}$.)}
\label{two_stage_scheme}
\end{figure}


Empirically, we find that the model architecture which achieves promising performance on the test dataset is the summation network of multi-layer perceptron (MLP) and multi-horizon auto-regressive model (MAR), denoted as ``MLP+MAR'', in both Stage 1 and Stage 2. Therefore, our model can be written as
\begin{equation}
    f^*(\vb{x}) = \vb{A}^{(0)}\vb{x} + \vb{b}^{(0)} + \text{MLP}^{(m)}(\vb{x}),
\end{equation}
\begin{equation}
    \text{MLP}^{(m)}(\vb{x}) = \phi^{(m)}( \vb{W}^{(m)}( \cdots \phi^{(1)}(\vb{W}^{(1)}\vb{x}+\vb{b}^{(1)})))+\vb{b}^{(m)}),
\end{equation}
where $m$ indicates the number of network layers in MLP, $\phi^{(*)}$ is the activation function. All parameters $\vb{W}^{(*)}$ and $\vb{b}^{(*)}$ in ``MLP+MAR'' model are jointly trained via stochastic gradient descent during the training phase.

Our designed hybrid framework MLP+MAR can model both non-linear and linear characteristics in $x_t$, via MLP and MAR respectively. Meanwhile, our model is optimized explicitly for capturing long-range time series dependencies in Stage 1. By using future time series $\vb{x}_{fut}$ to predict $\vb{x}_f$, it effectively utilizes the seasonal properties of $x_t$. Compared to SSL-based models which use random masking in pre-training stage, the two-stage model is more computationally efficient, and its Stage 1 results are more interpretable. In addition, it is an adaptable framework where one can use different model combinations in Stage 1 and Stage 2 to achieve the best performance.


%


\section{Experiments and Discussion}
In this section, we study the proposed two-stage algorithm empirically in comparison with other state-of-the-art forecasting algorithms on public datasets and discuss the insights from the proposed two-stage framework.

\subsection{Baseline Algorithms, Datasets and Metrics}
We compare the performance of our model against baseline algorithms, including multi-horizon auto-regressive model (MAR), multi-layer perceptron (MLP), stacked LSTM~\cite{hochreiter1997long}, SPAR~\cite{SPAR08}, previous period, as well as state-of-the-art seasonal time series forecast deep learning model RESTFul~\cite{wu2018restful}. Our two-stage framework is implemented in PyTorch and the main model configurations are listed in Table~\ref{2SModel}.

\begin{table}[!h]
\centering 
\caption{Two-stage network configuration with optimal forecast performance for $h=12$.}
\vspace{3mm}
\label{tab:simData}
\begin{threeparttable}
\begin{tabular}{|l|c|}
\hline
MLP Layers&  $[200,100,50]$  \\ \hline
Stage 1 Epochs &  $40$  \\ \hline
Stage 2 Epochs &  $20$  \\ \hline
Dropout &  $0.5$  \\ \hline
Learning Rate &  $0.01$  \\ \hline
Batch Size &  $64$  \\ \hline
Normalization &  Layer  \\ \hline
\end{tabular}
\end{threeparttable}
\label{2SModel}
\end{table}

We select M4 Competition Hourly datasets \cite{MAKRIDAKIS202054} as representative time series data to show the effectiveness of our proposed two-stage framework in multi-horizon forecasting. The M4 Hourly datasets consist of 414 seasonal time series and each has length between 700 to 960 points, collected from diverse fields including education, tourism, and services, etc.
We split each time series equally into two halves and use the first half for training and second for evaluation. We normalize each time series to zero mean and unit variance before training the forecast models and compute prediction accuracy with normalized values.

Given the true values of time series $x_t$ (with length $n$) and predicted values $\hat{x}_t$, we implement four error metrics to gauge the model's prediction accuracy.

\begin{equation}
\textrm{MAPE} = \frac{1}{n} \sum_{t=1}^n  \frac{|x_t-\hat{x}_t|}{|x_t|}
\end{equation}

\begin{equation}
\textrm{RMSPE} = \sqrt{\frac{1}{n} \sum_{t=1}^n  \frac{|x_t-\hat{x}_t|^2}{|x_t|^2}}
\end{equation}

\begin{equation}
\textrm{RMSE} = \sqrt{\frac{1}{n} \sum_{t=1}^n |x_t-\hat{x}_t|^2}
\end{equation}

\begin{equation}
\textrm{MAE} = \frac{1}{n} \sum_{t=1}^n |x_t-\hat{x}_t|
\end{equation}
In addition, we also compute each error metric after removing the worst 5\% predictions $\hat{x}_t$ from each time series (denoted as ``MAPE-95", etc.). This method reduces the prediction error due to possible anomalies in the time series.




\subsection{Forecasting Performance Comparisons}
We train the models to predict the time series in the forecast horizon of length $h=12$. We perform grid search to optimize the architecture of each model and experiment with different model combinations for Stage 1 and Stage 2. The results are summarized in Table \ref{PerfComp}. It can be observed that the MAR regression and MLP provide strong baselines, likely due to the low-dimensional and seasonal structure of $x_t$. The hybrid model MLP+MAR, which captures both linear and non-linear dependence in $x_t$, achieves the best baseline performance.  
RESTFul~\cite{wu2018restful}, the state-of-the-art forecast model of univariate seasonal time series, has sub-optimal accuracy. In the original work, the RESTFul encoder synthesizes the time series at different time scales (by averaging $x_t$ over previous day, week, and month) to predict the next day $(h=1)$. However, when predicting a longer horizon which is half of period length $(h=T/2)$, this encoding mechanism might not effectively capture the long-range structure in $(x_{t+1},...,x_{t+h})$. As a comparison, 
we can find that our proposed two-stage model outperforms the baseline methods in all but one error metric (RMSPE). This reveals that Stage 1 of our model has captured the long-range future behavior of the time series which is useful in predicting the forecast horizon in Stage 2.


\begin{table*}[!t]
\centering 
\caption{Performance of different horizon forecast algorithms on M4 Hourly dataset. Each model predicts a forecast horizon of length $h=12$. For each error metric, we report the prediction accuracy on the entire test set as well as excluding worst 5\% predictions of each time series to reduce the effects of possible anomalies in time series. The best results are highlighted.}
\vspace{1mm}
\label{tab:simData}
\begin{threeparttable}
\begin{tabular}{l||c|c|c|c|c|c|c|c}
\hline
Metric &  MAPE     &   MAPE-95    &  RMSPE    &  RMSPE-95 & RMSE   & RMSE-95 & MAE &  MAE-95  \\ \hline\hline

\textbf{Proposed Two-Stage}   &   \textbf{1.399}    &     \textbf{0.346}   & 11.629  &  \textbf{0.562} & \textbf{0.305} & \textbf{0.232} & \textbf{0.214} & \textbf{0.179} \\ \hline

MLP+MAR & 1.417 & 0.379 & \textbf{11.058} & 0.610 & 0.330 & 0.255 & 0.235 & 0.199 \\ \hline

MLP   &   1.423    &    0.410   &   11.197   &   0.605    &   0.385    &  0.305     &   0.281    &   0.242  \\ \hline

Deep-LSTM   &   1.539    &    0.459   &   11.857   &   0.652    &   0.422    &  0.341     &   0.320    &   0.278  \\ \hline

MAR   &   1.551    &    0.416   &   12.275   &   0.672    &   0.349    &  0.275     &   0.253    &   0.216  \\ \hline

RESTFul   &   1.642    &    0.451   &   11.808   &   0.721    &   0.375    &  0.301     &   0.276    &   0.238  \\ \hline

Previous Period   &   1.776    &    0.435   &   14.365   &   0.733    &   0.391    &  0.292     &   0.263    &   0.217  \\ \hline

SPAR-h12   &   2.077    &    0.570   &   15.825   &   0.869    &   0.447    &  0.364     &   0.340    &   0.297  \\ \hline

\end{tabular}\vspace{-0.4cm}
\end{threeparttable}
\label{PerfComp}
\end{table*}

\subsection{Ablation Study and Insights Discussion}
\subsubsection{Improving Baseline Models by Including Stage 1}

The key insight of the two-stage framework is that by learning the future horizon in Stage 1, the model will explicitly capture the long-range structure of the time series and use it to enhance the final prediction accuracy of Stage 2. To demonstrate the improvement in model performance resulting from Stage 1, we perform the control experiments where we use the current best baseline model in Stage 2 and augment it with another Stage 1 model. The Stage 1 output is directly concatenated to the input features of Stage 2. We compare the prediction accuracy of the original model with the two-stage augmented model.

The results are summarized in Table \ref{ModelAug}. We find that for baseline models with different forecast horizons $h\in \{6,12,24\}$, the augmented two-stage model consistently outperforms the baseline in all but one error metric (RMSPE). In addition, the improvement is robust against perturbations to the detailed architecture of Stage 1 models. This shows that the two-stage framework is an effective model-augmentation technique to enhance the prediction accuracy.

\begin{table*}[!t]
\centering 
\caption{Improving prediction accuracy by including Stage 1 in baseline models with forecast horizon $h\in \{6,12,24\}$. “BL” means the best baseline model (MLP+MAR) and “2S” means two-stage. The best results for each $h$ are highlighted.}
\vspace{1mm}
\label{tab:simData}
\begin{threeparttable}
\begin{tabular}{c|l||c|c|c|c|c|c|c|c}
\hline
Horizon $h$ & Model &  MAPE     &   MAPE-95    &  RMSPE    &  RMSPE-95 & RMSE   & RMSE-95 & MAE &  MAE-95  \\ \hline\hline

6 & BL & 1.265 & 0.332 & \textbf{10.008} & 0.542 & 0.285 & 0.219 & 0.201 & 0.169 \\ \hline

6 & 2S &    \textbf{1.214}    &     \textbf{0.305}   & 10.102  &  \textbf{0.496} & \textbf{0.266} & \textbf{0.201} & \textbf{0.185} & \textbf{0.155} \\ \hline

12 & BL & 1.454 & 0.384 & \textbf{11.423} & 0.617 & 0.331 & 0.258 & 0.237 & 0.201 \\ \hline

12 & 2S &    \textbf{1.399}    &     \textbf{0.346}   & 11.629  &  \textbf{0.562} & \textbf{0.305} & \textbf{0.232} & \textbf{0.214} & \textbf{0.179} \\ \hline

24 & BL & 1.511 & 0.405 & \textbf{11.833} & 0.651 & 0.349 & 0.273 & 0.251 & 0.214 \\ \hline

24 & 2S &    \textbf{1.489}    &     \textbf{0.374}   & 11.900  &  \textbf{0.614} & \textbf{0.319} & \textbf{0.247} & \textbf{0.226} & \textbf{0.191} \\ \hline

\end{tabular}\vspace{-0.4cm}
\end{threeparttable}
\label{ModelAug}
\end{table*}

\begin{table*}[!t]
\centering 
\caption{Performance of two-stage models with different future horizons $H$ (fixing $h=12$). The optimal prediction accuracy (highlighted) occurs at $H^*=12$, where MSE of Stage 1 model is also the lowest.  The best results are highlighted.}
\vspace{1mm}
\label{tab:simData}
\begin{threeparttable}
\begin{tabular}{c||c|c|c|c|c|c|c|c||c}
\hline
$H$ &  MAPE     &   MAPE-95    &  RMSPE    &  RMSPE-95 & RMSE   & RMSE-95 & MAE &  MAE-95 & S1 MSE  \\ \hline\hline

0 & 1.454 & 0.384 & \textbf{11.423} & 0.617 & 0.331 & 0.258 & 0.237 & 0.201 & N/A  \\ \hline

6 & 1.417 & 0.351 & 11.715 & 0.570 & 0.308 & 0.235 & 0.216 & 0.182 & 0.151  \\ \hline

12   &   \textbf{1.399}    &     \textbf{0.346}   & 11.629  &  \textbf{0.562} & \textbf{0.305} & \textbf{0.232} & \textbf{0.214} & \textbf{0.179} & \textbf{0.147} \\ \hline

18 & 1.462 & 0.356 & 12.052 & 0.582 &  0.309 & 0.237 & 0.218 & 0.183 & 0.158 \\ \hline

24 & 1.470 & 0.356 & 11.985 & 0.585 & 0.306 & 0.234 & 0.215 & 0.180 & 0.166  \\ \hline

\end{tabular}\vspace{-0.4cm}
\end{threeparttable}
\label{TuneH}
\end{table*}

\subsubsection{Tradeoff Between Future Horizon Length}

A critical hyperparameter of the two-stage model is the future horizon length $H$. If we choose a larger $H$, Stage 1 model captures long-range future behavior of the time series which would help predict the immediate forecast horizon. On the other hand, the longer future horizon is also more difficult to predict in Stage 1, reducing the accuracy of inputs to Stage 2. Therefore, one expects that there exists an optimal future horizon length $H^*$ which could balance this tradeoff between long-range future behavior and smaller forecasting error in Stage 1. 

To rigorously analyze this tradeoff, we perform a control experiment. We fix the Stage 1 and Stage 2 model architectures (with $h=12$) and vary the length of future horizon $H \in \{0,6,12,18,24 \}$. The results are summarized in Table \ref{TuneH}. Indeed, we find that the prediction accuracy of the two-stage model is sensitive to the choice of $H$ and the optimal performance occurs at $H^*=12$. In addition, we observe that Stage 1 evaluation MSE (last column of Table \ref{TuneH}) is a good indicator of the final performance of Stage 2. This confirms that the future horizon length $H$ is a critical hyperparameter which needs to be carefully selected in order to obtain desirable forecasting performance.

\section{Conclusions and future directions}\label{sec:conc}
\vspace{-0.2cm}
We propose a novel two-stage framework, based on SSL, to forecast univariate time series with seasonality. Our model efficiently captures the long-range time series structure in Stage 1 and utilizes future time series to predict the forecast horizon in Stage 2. Furthermore, 
we incorporate auto-regressive model and neural network model to for linear and non-linear characteristics in time series, respectively, to improve the forecasting accuracy.
We demonstrate that our model achieves state-of-the-art performance on M4 Hourly dataset and can be applied as a model-augmentation technique to enhance the prediction accuracy of existing models. 

Currently, Stage 1 and Stage 2 models in our framework are trained independently. To better synthesize the predictions of the two models, we can jointly update $f_1$ and $f_2$ in the training step of Stage 2. Furthermore, we can apply randomized training in Stage 2 where $f_2$ receives the true values of future horizon $\vb{x}_{fut}$ with probability $p$ and Stage 1 predictions $f_1$ ($\vb{x}_{his}$) with probability $1-p$. These techniques are left as future work to further improve the prediction accuracy.


\bibliographystyle{IEEEbib}
\bibliography{5_bibfile}

\end{document}